\documentclass[runningheads]{llncs}

\usepackage{eccv}

\usepackage{eccvabbrv}

\usepackage{graphicx}
\usepackage{booktabs}

\usepackage[accsupp]{axessibility}  

\usepackage{hyperref}

\usepackage{orcidlink}

\usepackage{booktabs}
\usepackage{colortbl}
\usepackage{url}
\usepackage{wrapfig}
\usepackage{blindtext}
 \usepackage{multirow}
\usepackage[utf8]{inputenc} 
\usepackage[T1]{fontenc}    
\usepackage{amsfonts}       
\usepackage{nicefrac}       
\usepackage{microtype}      
\usepackage{xcolor}         
\usepackage{bbding}
\usepackage{soul}
\usepackage{changes}

\newcommand{\red}[1]{\textcolor[rgb]{1,0,0}{#1}}

\newcommand{\green}[1]{\textcolor[rgb]{0,0.69,0.31}{#1}}
\definecolor{officegreen}{rgb}{0.0, 0.5, 0.0}

\begin{document}

\title{NL2Contact: Natural Language Guided 3D Hand-Object Contact Modeling with \\ Diffusion Model} 

\titlerunning{NL2Contact}

\author{Zhongqun Zhang\inst{1}
\and Hengfei Wang\inst{1}
\and Ziwei Yu\inst{2}
\and Yihua Cheng\inst{1}\thanks{Corresponding author.}
\and \\ Angela Yao\inst{2}
\and Hyung Jin Chang\inst{1}
}

\authorrunning{Z.~Zhang et al.}

\institute{University of Birmingham, UK \and
National University of Singapore, Singapore
}

\maketitle
\begin{abstract}
Modeling the physical contacts between the hand and object is standard for refining inaccurate hand poses and generating novel human grasp in 3D hand-object reconstruction.
However, existing methods rely on geometric constraints that cannot be specified or controlled.
This paper introduces a novel task of controllable 3D hand-object contact modeling with natural language descriptions. Challenges include
i) the complexity of cross-modal modeling from language to contact, and ii) a lack of descriptive text for contact patterns.
To address these issues, we propose NL2Contact, a model that generates controllable contacts by leveraging staged diffusion models. Given a language description of the hand and contact, NL2Contact generates realistic and faithful 3D hand-object contacts. To train the model, we build \textit{ContactDescribe}, the first dataset with hand-centered contact descriptions. It contains multi-level and diverse descriptions generated by large language models based on carefully designed prompts (e.g., grasp action, grasp type, contact location, free finger status). We show applications of our model to grasp pose optimization and novel human grasp generation, both based on a textual contact description.
\end{abstract}
\section{Introduction}
\label{sec:intro}

Understanding and modeling physical contact between hands and objects~\cite{brahmbhatt2019contactdb, brahmbhatt2020contactpose, taheri2020grab, xie2023nonrigid, yang2021cpf} can contribute to 
applications in animation~\cite{lakshmipathy2023contact}, virtual reality~\cite{wang2023high}, augmented reality~\cite{wu2022saga, tendulkar2023flex}, and robotics~\cite{zhang2021manipnet, qin2022dexmv}.
The contact could also improve hand-object 3D pose and shape estimation, either reconstructed from images~\cite{grady2021contactopt, zhou2022toch, tse2022s} or generated from given object models~\cite{li2022contact2grasp, jiang2021hand, liu2023contactgen}. 

\begin{figure*}[t]
\centering
\includegraphics[width=1.0\linewidth]{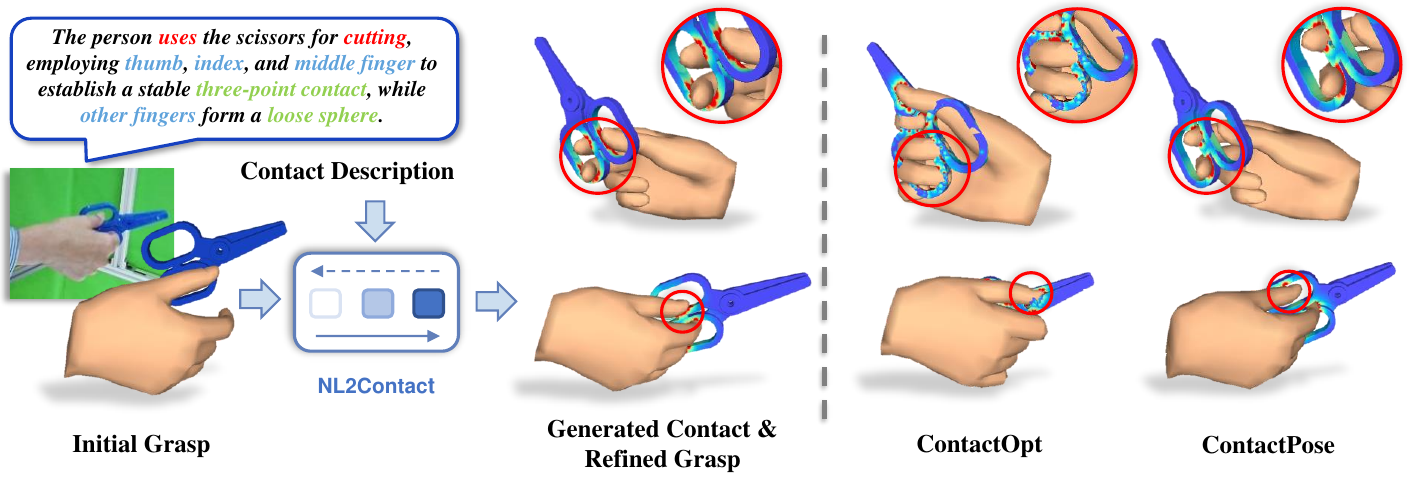}
\caption{
\textbf{Illustration of the NL2Contact setting.} Our method learns to model 3D hand-object contact from language description. The modeling contact can be used to grasp pose optimization. By utilizing the predicted contact to refine the initial grasp, our performance surpasses that of ContactOpt~\cite{grady2021contactopt} and ContactPose~\cite{brahmbhatt2020contactpose}. It's worth noting that ContactPose is the annotation from the motion capture system. 
}

\label{fig:highlevel_framework}
\end{figure*}

Conventional contact modeling methods typically apply PointNet~\cite{qi2017pointnet} on hand-object point clouds to infer the contact status of each point. 
These methods leverage geometric constraints for modeling physically realistic contact.
However, 
their results do not align well with human grasps~\cite{liu2023contactgen}.  
As shown in Figure~\ref{fig:highlevel_framework}, the method ContactOpt~\cite{grady2021contactopt}, despite being state-of-the-art, produces a result in which all the fingers make contact with scissors.  While mostly feasible physically, this is unrealistic and does not correspond to 
human usage of scissors.  

Recently, there has been growing interest in synthesizing natural hand-object interactions based on object semantics and affordance, as well as human behavior and intent. 
Existing methods typically use intent labels, such as interaction verbs~\cite{liu2022hoi4d, yang2022oakink} and object part-level affordance~\cite{jian2023affordpose}, to achieve controllable hand-object interaction modeling.
Interaction verbs, like \textit{use} and \textit{handle}~\cite{yang2022oakink}, describe the behavior and intent during the interaction but are typically high-level and vague to guide precise contact modeling. Affordance, labeled on 3D point clouds of objects, indicates feasible regions where interactions occur.  However, the regions are coarse and require prior knowledge of object categories. 
These methods face significant challenges in precisely controlling hand-object contact modeling.

This paper introduces a novel task, natural language-guided 3D hand-object contact modeling.
Inputting a natural language description of hand-object interaction, our task outputs physically realistic contacts aligned with the description, as shown in Figure~\ref{fig:highlevel_framework}.
Compared to verbs and affordances, natural language descriptions provide freer and more precise expressions of contact patterns. To this end, we propose the \textit{ContactDescribe} dataset, which provides 3D hand-object contact and fine-grained natural language descriptions. To our knowledge, this is the first dataset providing fine-grained descriptions for hand-object contact.

We create \textit{ContactDescribe} using a novel text annotation scheme. Initially, we designed hand-centered prompts. 
Subsequently, we aim to annotate detailed natural language descriptions based on these prompts. However, manual annotation of text descriptions is time-consuming, and annotator expertise significantly influences the results. In this work, we propose leveraging large-scale language models (LLMs)~\cite{kojima2022large} for annotation. We carefully craft an input containing our prompts and let the power of LLM provide descriptions; 
with well-selected prompts, the resulting descriptions are diverse and fine-grained.

We also propose a new cross-modal latent diffusion model for our task. Directly modeling hand-object contacts from text descriptions is non-trivial,  
since natural language descriptions do not capture the precise physical 3D interaction between hand and object.
As such, we design a two-stage approach. Initially, we generate a hand pose from a noisy hand input and text description, where the hand pose aligns with the text description but may not establish optimal contact with the object. We then further model the hand-object contact from the text description and the object's point clouds, with the hand pose serving as a prompt. To our knowledge, we are the first to model 3D hand-object contacts using natural language descriptions.

In summary, our contributions are four-fold:
\begin{itemize}
    \setlength{\itemsep}{0pt}%
    \setlength{\parskip}{0pt}%
    
    \item We explore a new task of modeling 3D hand-object contact from natural descriptive languages.
    
    \item We create a new dataset, \textit{ContactDescribe}, consisting of contact maps and corresponding descriptive languages generated by LLMs. 

    \item We design a new staged diffusion model to generate 3D hand-object contacts from the descriptive language. 

    \item We propose a text-to-hand-object fusion network that efficiently extracts contact information from point clouds and natural language to guide the reverse diffusion process. 
\end{itemize}
We evaluate the efficacy of our approach on two datasets: the newly introduced \textit{ContactDescribe} dataset and the HO3D dataset~\cite{hampali2020honnotate}. Additionally, we validate the capability of our methods to effectively capture controllable 3D hand-object contact through our natural language annotations. 
Our dataset and proposed method provide a new research direction for the community to model interactions with natural language.  
\section{Related Work}

\noindent\textbf{Hand-object contact modeling.}
Modeling contact between hands and objects is important for grasp reconstruction and synthesis~\cite{tse2022collaborative, tse2023spectral, hasson20_handobjectconsist, liu2021semi, sener2022assembly101}. Existing studies~\cite{grady2021contactopt, li2022contact2grasp, xie2023nonrigid, zhou2022toch, tse2022s, jiang2021hand, yang2021cpf, yu2022uv, zhu2023contactart} inferred contact patterns from hand-object point clouds with geometry constraints such as penetrations. ContactOpt~\cite{grady2021contactopt}, GraspTTA~\cite{jiang2021hand}, and Contact2Grasp~\cite{li2022contact2grasp} generated the contact regions based on PointNet~\cite{qi2017pointnet}. ContactGen~\cite{liu2023contactgen} further generated contact location with the hand part and touch direction. Additional constraints from RGB images and sequences also benefit the contact modeling. S$^2$Contact~\cite{tse2022s} proposed estimated the contact with visual and geometry constraints. TOCH~\cite{zhou2022toch} introduces the temporal constraint to model contact sequences. 
However, research focusing on text-to-contact modeling remains relatively limited.

\noindent\textbf{Text-guided diffusion models.}
Diffusion models~\cite{ho2020denoising, liu2022compositional} are probabilistic generative models that use the Markov chain to convert a simple Gaussian distribution to a target data distribution by gradually adding and removing noise. Recently, Rombach~\etal~\cite{rombach2022high} proposed the latent diffusion model (LDM) and has shown the real powers of LDM in text-to-image synthesis~\cite{wang2024textgaze}. Inspired by LDM, Ye~\etal~\cite{ye2023diffusion} also adopt a text-conditioned diffusion model for
guiding the 3D generation of hand-object interactions. Furthermore, text is often used in human motion generations~\cite{kong2023priority, ahn2018text2action, guo2022generating}. Karunratanakul~\etal~\cite{karunratanakul2023guided} proposed the Guided Motion Diffusion, which can generate controllable and diverse human motions given a text prompt. Text inputs also show potential improvements in robotic grasping~\cite{tang2023graspgpt, cheang2022learning}. Ha~\etal~\cite{ha2023scaling} presented a framework for robotic skill learning, which uses natural language (\emph{e.g.} put the toy into the left bin) to guide high-level grasp planning. This work uses the diffusion model to explore the controllable 3D hand-object contact generation from text description.
\section{\textit{ContactDescribe} Dataset} 
\label{sec:contactdescribe}

\begin{figure}[t]
\centering
  \begin{minipage}{0.56\linewidth}
   \includegraphics[width=\linewidth]{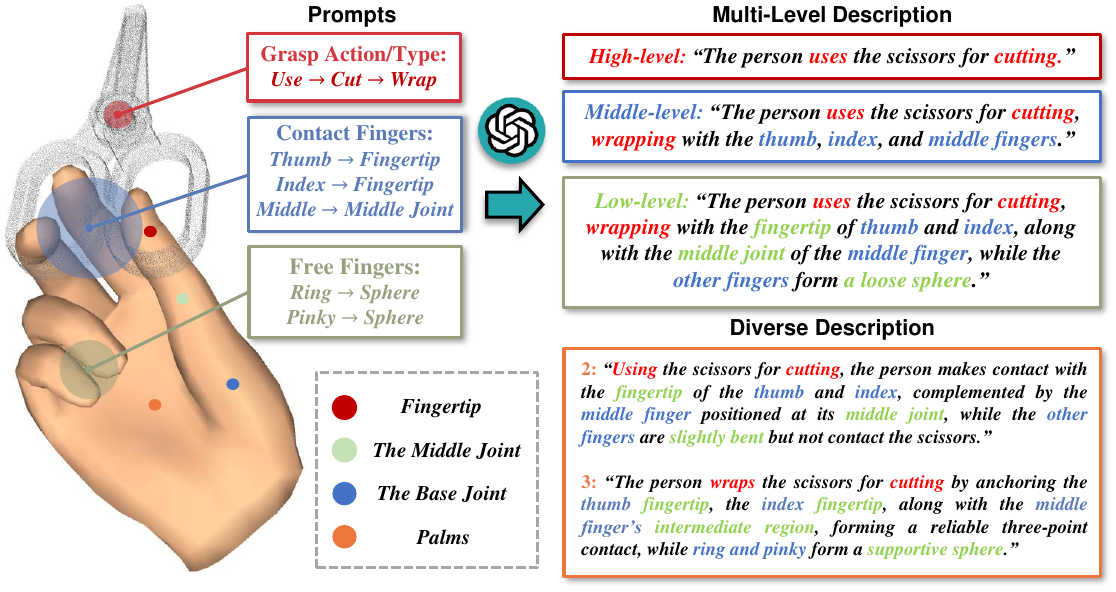}
\label{fig:attribute_plot}
\caption{An example of the dataset with multi-level
language descriptions.
Given the proposed hand-centered contact prompts, our dataset leverages ChatGPT3~\cite{chatgpt} to generate multi-level and diverse free-form text descriptions of contact patterns. 
}
\label{fig:dataset}
\end{minipage}
~  
\begin{minipage}{0.4\linewidth}
\centering
\resizebox{\linewidth}{!}{

\begin{tabular}{l|cc|ccc}
    \toprule[1pt]
    \multirow{2}{*}{\textbf{Datasets}}
    &\multicolumn{2}{c|}{\textbf{Label}}
    & \multirow{2}{*}{\shortstack{\textbf{Grasp}\\ \textbf{type}}} & \multirow{2}{*}{\shortstack{\textbf{Finger}\\ \textbf{status}}} & \multirow{2}{*}{\shortstack{\textbf{Contact}\\ \textbf{location}}}
    
    \\
 
     & Verb& Sentence&  & &\\
     
    \hline
    HO3D~\cite{hampali2020honnotate} & \red{\XSolidBrush} & \red{\XSolidBrush}  &\red{\XSolidBrush}  &\red{\XSolidBrush} &\red{\XSolidBrush} \\
   
    YCBAfford~\cite{corona2020ganhand} & \red{\XSolidBrush} & \red{\XSolidBrush}   &\green\Checkmark  &\red{\XSolidBrush} &\red{\XSolidBrush} \\
    
   
    AffordPose~\cite{jian2023affordpose} & \red{\XSolidBrush} & \red{\XSolidBrush}   &\green\Checkmark  &\red{\XSolidBrush} &\red{\XSolidBrush} \\
    
    HOI4D~\cite{liu2022hoi4d} & \green\Checkmark & \red{\XSolidBrush}   &\green\Checkmark &\red{\XSolidBrush} &\red{\XSolidBrush}  \\
    ContactPose~\cite{brahmbhatt2020contactpose} & \green\Checkmark & \red{\XSolidBrush}  &\green\Checkmark  &\red{\XSolidBrush} &\red{\XSolidBrush}  \\
    OakInk~\cite{yang2022oakink} & \green\Checkmark & \red{\XSolidBrush}  &\green\Checkmark  &\red{\XSolidBrush} &\red{\XSolidBrush}  \\
    
    \rowcolor[rgb]{0.902,0.902,0.902}\textbf{Ours}  & \green\Checkmark & \green\Checkmark &\green\Checkmark &\green\Checkmark &\green\Checkmark \\
    \bottomrule[1pt]
\end{tabular}}

\captionof{table}{Intent labels comparison between different
datasets of hand-object interaction. Most existing datasets collect verb-level intent labels. They are insufficient to prompt controllable contact modeling. Our dataset is the first to provide detailed sentence-level contact descriptions.}
\label{table:annotations}
\end{minipage}
\end{figure}

\noindent\textbf{Dataset collection.} Learning a controllable 3D hand-object contact modeling from natural language requires 
3D contact data with
corresponding text descriptions.
No dataset to date includes detailed text for describing hand-object contact patterns (see comparison in Table~\ref{table:annotations}). To bridge this gap, we
created the \textit{ContactDescribe} dataset, which contains paired point cloud-text data. The hand-object point clouds and essential annotations (such as 
MANO~\cite{romero2017embodied} parameters, object meshes and 6D poses, and hand-object contact maps.) are taken from ContactPose~\cite{brahmbhatt2020contactpose}. 
\textit{ContactDescribe} 
extends the ContactPose dataset by adding  
annotations of contact pattern descriptions for each hand-object point cloud. To enhance the naturalness and diversity of these descriptions, we leverage Large Language Models (LLMs), such as ChatGPT~\cite{chatgpt}, to generate the final sentences given hand-centered contact prompts, as depicted in Figure~\ref{fig:dataset}.

\noindent\textbf{Multi-level language description.} 
As illustrated in Figure~\ref{fig:dataset}, we newly designed a coarse-to-fine multi-level text description to describe the hand-object contact systematically. The high-level text describes the grasp action, the middle-level text details the grasp type and finger status, while the low-level text specifies the contact location on which hand joints.

\noindent\textbf{Prompt annotation.}
We first manually annotate each hand-object point cloud to obtain the contact prompts.
We aim to depict contact from a hand-centric perspective and prioritize descriptions that enhance adaptability across diverse object categories and attributes. 

As illustrated in Figure~\ref{fig:dataset}, our prompt annotations of the contact parts contain several single-choice questions covering grasp action, basic grasp type, contact location on each finger, and free finger status. The basic grasp types assume the involvement of all fingers in contact, including \emph{`wrap', `pinch', `lift',} and \emph{`twist'}. However, 
grasp type alone cannot precisely describe contact patterns and lacks human-intuitive {understanding}. 
Moreover, the diverse vocabulary used to describe grasps presents challenges in achieving uniform descriptions. Consequently, we ask annotators to specify the contacts of each finger to refine the depiction. Three descriptive options are provided for fingers with free contact: \emph{`sphere', `full extension'}, and \emph{`partial extension'.} To make our language more natural and be used by untrained users in daily chat, we avoid using highly specialized vocabulary to describe hand joints, such as the Distal Interphalangeal Joint (DIP Joint).

\noindent\textbf{Sentence generation.}
To increase description diversity, we leverage the ChatGPT3 model~\cite{chatgpt} for sentence generation and provide the following prompts:

\begin{itemize}
\scriptsize
    \setlength{\itemsep}{0pt}%
    \setlength{\parskip}{0pt}%
    \item[1.] \textit{I will give you some prompts describing a hand gesture when humans use the scissors; please help me reword a sentence. But remember not to describe the shape or attribute of the object. }
    \item[2.] \textit{Prompts are [grasp action], [grasp type], [contact location], [free fingers status].}
    \item[3.] \textit{To tell ChatGPT some basic knowledge about the hand, \emph{e.g.} using the fingertip to represent the Proximal/DIP joint, what the sphere finger is, etc.}
    \item[4.] \textit{Help me reword four sentences to a different format but keep their meaning. You can incorporate the descriptions I just provided.}
\end{itemize}
We further leverage ChatGPT to check the sentence and give feedback to annotators automatically, and annotators rephrase inappropriate sentences manually. 

\noindent\textbf{Detailed Statistics.}
\textit{ContactDescribe} contains 2,300 unique grasps of 25 household objects (e.g. scissors, bottles, eyeglasses, \etc.) from 50 participants. Our dataset consists of 11,500 language descriptions, with each grasp corresponding to five different sentences.

\section{Methodology}
\subsection{Preliminary: Contact Modeling}
 The contact map on hand point cloud $C_H\in \mathbb{R}^{N \times 1}$ and object point cloud $C_O \in \mathbb{R}^{N \times 1}$, where $N$ is the number of points, denotes the contact probability of each point, with values constrained within the [0, 1] range. As shown in Figure~\ref{fig:highlevel_framework} and~\ref{fig:Pipeline}, the redder the color, the higher the probability of the point being in contact. The contact modeling network is applied to the point cloud and outputs its contact map.
Our goal is to model the contact map from the input text description. The contact map can be easily applied for the following two tasks:

\noindent\textbf{Grasp pose optimization} leverages contact map to refine inaccurate 3D hand pose, e.g. the estimation from a single RGB image. It first predicts the contact map on both hand and object mesh. To optimize the hand pose, the hand MANO~\cite{romero2017embodied} parameters are iteratively
updated to minimize the difference
between the current contact maps and the predicted contact maps.

\noindent\textbf{Human grasp generation} generates hand parameters as human grasps from a given object mesh based on the contact map. It first generates hand MANO mesh and then predicts the contact map between the object and the generated hand. Finally, the hand-object meshes are fitted using the contact map.

\subsection{Method Overview}
\begin{figure*}[t]
\centering
\includegraphics[width=1.0\linewidth]{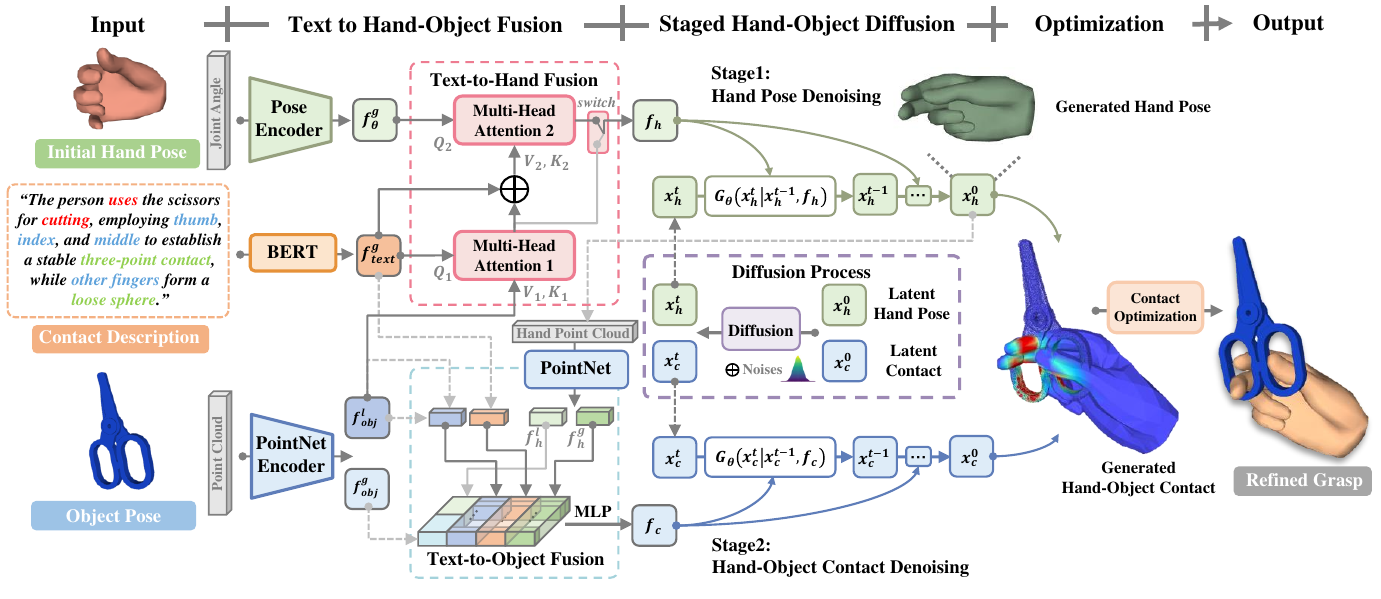}
\caption{
\textbf{NL2Contact pipeline.} We propose a novel method to model 3D hand-object contact using language description. Our framework is composed of 1) a Text-to-Hand-Object Fusion module, efficiently fusing the geometry information from both point clouds and semantic information from natural language, 2) a staged latent diffusion model to first generate the hand pose fitting the hand description and then generate the 3D hand-object contact map conditioned on the descriptions and the generated hand pose. 3) contact optimization, which iteratively optimizes the hand pose using the generated contact. We also added a switch to toggle between its application for grasp pose optimization and human grasp generation.}
\label{fig:Pipeline}
\end{figure*}

Given inputs $\{\widetilde{\mathcal{H}}, \mathcal{O}, \mathbb{T}\}$, where $\mathcal{O} \in \mathbb{R}^{2048 \times 3}$ and $\mathbb{T}$ are the object point cloud and the contact description respectively. The initial hand MANO pose $\widetilde{\mathcal{H}}=\left(\theta, \beta, t^H, R^H\right)$ consisting of pose, shape, translation, and rotation respectively is an optional input. In this work, we consider initial hand poses derived from an image-based algorithm~\cite{hasson20_handobjectconsist}, though our method is not limited specifically and can accept alternatively derived hand poses. We aim to generate a 3D contact map $\hat{C}_H$ and $\hat{C}_O$ on both the hand and object mesh. 
To achieve this, a three-module generative network, NL2Contact, is proposed as shown in Figure~\ref{fig:Pipeline}. 
The text-to-hand-object fusion module (Section~\ref{sec:fusion}) efficiently extracts contact information from both point clouds and natural language to guide the reverse diffusion process. A staged diffusion module (Section~\ref{sec:staged diffusion}) generates hand MANO pose and subsequently predicts 3D hand-object contact maps. The generated hand pose is then refined through optimization using the generated contact map (Section~\ref{sec:optimization}). We elaborate on these submodules in the following sections.

\subsection{Text to Hand-Object Fusion}
\label{sec:fusion}
\noindent\textbf{Encoders.} Consider a set of inputs $\{\widetilde{\mathcal{H}}, \mathcal{O}, \mathbb{T}\}$, the NL2Contact extracts global feature $f_\theta^g \in \mathbb{R}^{64}$ and tokens embedding $f_{text} \in \mathbb{R}^{768 \times n}$ from the hand pose $\widetilde{\mathcal{H}}$ and text $\mathbb{T}$ by two feature encoders VPoser~\cite{pavlakos2019expressive} and BERT~\cite{devlin2018bert}, and extracts global and local features $f_{obj}^g \in \mathbb{R}^{64 \times N}$ and $f_{obj}^l \in \mathbb{R}^{1024}$ from object point clouds $\mathcal{O}$ using the PointNet++~\cite{qi2017pointnet++}, where $N$ is the number of points and $n$ is the number of words in text.

\noindent\textbf{Text-to-hand fusion.}
We apply MLP layers to the tokens embedding $f_{text}$ to get a global text feature $f_{text}^g \in \mathbb{R}^{64}$. Then we fuse $f_{text}^g$ with the hand pose feature $f_{\theta}^g$ 
and object global feature $f_{obj}^g$ via two multi-head attention modules
~\cite{carion2020end} in a cascaded manner to get the latent condition for the hand pose denoising process. 
The first multi-head attention module takes the text feature $f_{text}^g$ as a query to extract the object shape information from the object global feature $f_{obj}^g$. The second attention takes the hand pose feature $f_{\theta}^g$ as a query to extract the semantic hand gesture information from the text feature $f_{text}^g$. Such design allows a switch between grasp pose optimization and human grasp generation.
 
\noindent\textbf{Text-to-object fusion.} This module aims to fuse text information into hand-object point clouds to get the latent condition for guiding the hand-object contact denoising process. In particular, we first transfer the generated hand pose $\mathcal{H}^{\prime}$, which outputs from the first diffusion stage to the hand point by a differentiable MANO Layer~\cite{romero2017embodied}. Then, we extract  global and local features $f_{h}^g \in \mathbb{R}^{64 \times 778}$ and $f_{h}^l \in \mathbb{R}^{1024}$ from hand point clouds using the PointNet++~\cite{qi2017pointnet++}. To obtain the per-point hand-object feature, we combine the hand local feature $f_{h}^l$ and object local feature $f_{obj}^l$ together, including an additional binary per-point feature indicating whether the point belongs to the hand or the object. The global features of hand $f_{h}^g$, object $f_{obj}^g$ and text $f_{text}^g$ are then duplicated $N$ times, $N$ is (2048+778), for matching the shape of hand-object per-point feature. All these features are then concatenated and fused by MLP layers to produce the conditional embeddings $f_c$ for the contact generator below.

\subsection{Staged Hand-Object Diffusion Models.}
\label{sec:staged diffusion}
\noindent\textbf{Conditional latent diffusion models.}
Latent diffusion models (LDMs)~\cite{rombach2022high}
learns a data distribution $q\left(x^0\right)$ by reversing a progressive noise diffusion process from a prior Gaussian distribution $\mathcal{N}\left(\mu, \sigma^2\right)$. Given a sample $x^0 \sim q\left(x^0\right)$ from the latent space, a fixed diffusion process $q\left(x^{1: t} \mid x^0\right)=\prod_{t=1}^t q\left(x^t \mid x^{t-1}\right)$ 
is used to add Gaussian noise $q\left(\mathrm{x}^t \mid \mathrm{x}^{t-1}\right):= \mathcal{N}\left(\sqrt{1-\beta_t} \mathrm{x}^{t-1}, \beta_t \mathbf{I}\right)$, producing a chain of noisy latent codes $\left\{x^1, x^2, \ldots, x^t\right\}$, where $\beta_t$ is the 
variance defined in the forward process.
The latent code $x^0$ is extracted using a pre-trained Encoder~\cite{van2017neural} from the raw data, which reduces the computational requirements by learning this distribution in a latent space.
Then, the denoiser $G_\theta\left(\mathrm{x}^{t-1} \mid \mathrm{x}^t\right)=\mathcal{N}\left(\mu_\theta\left(\mathrm{x}^t, t\right), \sigma_t^2 \mathbf{I}\right)$\footnote{We slightly abuse the notion $\theta$ to represent learning of model parameters.}, typically a U-Net~\cite{ronneberger2015u}, is trained to estimate the noise from those samples. The training objective can be simplified to:
\begin{equation}
\begin{aligned}
& \mathcal{L}_{L D M}=\mathbb{E}_{\mathbf{x}^0, \varepsilon \sim \mathcal{N}(0,1), t}\left[\left\|\varepsilon-G_\theta\left(\mathbf{x}^t, t, f\right)\right\|_2^2\right], \\
& G_\theta\left(\mathrm{x}^{0: t}\right)=G\left(\mathrm{x}^t\right) \prod_{t=1}^t G_\theta\left(\mathrm{x}^{t-1} \mid \mathrm{x}^t, f\right), \\
\end{aligned}
\end{equation}
where $f$ is the latent codes of conditions, such as text and images. \\

\noindent\textbf{Hand pose diffusion.}
Figure~\ref{fig:Pipeline} shows that we use LDMs conditioned on Text-to-Hand embeddings $f_h$ to model the hand pose generation process. Following the LDMs, we apply encoder $E_p$ to encode the hand pose ground truth $\mathcal{H}_{gt}$ into a latent code and reshape it to dimensions $ x_h^0 \in \mathbb{R}^{d \times d}$.  As $E_p$, we use VPoser~\cite{pavlakos2019expressive}, a variational autoencoder for human pose representations.
Then, a forward diffusion process adds Gaussian noise to the input $\mathbf{x}^0$ to transfer it to $\mathbf{x}^t$. This latent code is subsequently fed into a denoiser $G^1_\theta$, U-Net~\cite{ronneberger2015u}, along with condition embedding $f_h$ and $t$ to predict the $\hat{\mathbf{x}}^0_h=G_{\theta}^1\left(\left[\mathbf{x}^t_h, f_h, t\right]\right)$, where $\mathbf{x}^0=E_p\left(\mathcal{H}_{gt}\right)$.
Finally, the decoder from VPoser generates the hand pose $\mathcal{H}^{\prime}$ from the reconstructed $\hat{\mathbf{x}}^0_h$. For the denoising process at the timestep $t$, the training objective minimizes the following:
\begin{equation}
\mathcal{L}_{\text {pose}}=\mathbb{E}_{\mathbf{x}^0_h, \varepsilon \sim \mathcal{N}(0,1), t}\left[\left\|\varepsilon-G^1_\theta\left(\mathbf{x}^t_h \mid t, f_h\right)\right\|_2^2\right].
\end{equation}

\noindent\textbf{Hand-object contact diffusion.}
We freeze the hand pose diffusion's parameters and 
pre-train a PointNet~\cite{qi2017pointnet++} as the contact encoder to extract the latent code from the hand-object contact groud-truth $\mathcal{C}_{gt}$ and then reshape it to $ x_c^0 \in \mathbb{R}^{32 \times 32}$. We get the condition embedding $f_c$ by sending the generated hand pose into the Text-to-object fusion module. Similar to the hand pose denoising, this embedding is then fed into the U-Net denoiser $G^2_\theta$ along with $x_c^t$ and $t$ to predict the $\hat{\mathbf{x}}^0_c$. The loss function is:
\begin{equation}
\mathcal{L}_{\text {contact}}=\mathbb{E}_{\mathbf{x}^0_c, \varepsilon \sim \mathcal{N}(0,1), t}\left[\left\|\varepsilon-G^2_\theta\left(\mathbf{x}^t_c \mid t, f_c\right)\right\|_2^2\right].
\end{equation}
The PointNet contact decoder produces the contact map $\hat{\mathcal{C}}$ from the predicted $\hat{\mathbf{x}}^0_h$.
At test time, a latent vector is randomly sampled from the Gaussian distribution and progressively denoised by U-Net to generate contact.

\subsection{Contact Optimizations.}
\label{sec:optimization}
Given the generated hand pose $\mathcal{H}^{\prime}$, our target is to optimize it to find the best parameter $\theta$ and $\beta$ that best complies with the generated hand-object contact map $\hat{\mathcal{C}}$. To achieve this, we iteratively minimize the difference between the current contact $\mathcal{C}^\prime$ and the generated contact $\hat{\mathcal{C}}$ and update the hand pose parameters.
We applied the contact map on both hand and object point clouds, where values are
bounded within the [0, 1] range to indicate the contact status of the points.
The current hand-object contact map $\mathcal{C}^\prime$ at each step is computed by DiffContact~\cite{grady2021contactopt}. The loss for the
contact optimization is formulated as follows:
\begin{equation}
\mathcal{L} _{opt} = \left\|\mathcal{C}^\prime_H - \hat{\mathcal{C}}_H \right\|_2^2 + \lambda_{o} \left\|\mathcal{C}^\prime_O - \hat{\mathcal{C}}_O \right\|_2^2 +\lambda_{pen} \mathcal{L}_{pen}, 
\end{equation}
where $\mathcal{L}_{pen}$ is the penetration loss to penalize
the penetration between the hand and the object, similar to ContactOpt~\cite{grady2021contactopt}.
\section{Experiments}

\subsection{Implementation Details}
Our model is trained in an end-to-end manner. A PointNet++~\cite{qi2017pointnet++} is pre-trained on the ContactPose~\cite{brahmbhatt2020contactpose} dataset as the contact encoder.
We sample 2048 points from the object mesh as the input. 
 We adopt the Adam optimizer~\cite{kingma2014adam}, a
batch size of 64, and a learning rate of $ 1.5 e^{-3}$ with 50 epochs for training hand pose denoising and another 100 epochs for hand-object contact generation. The training takes about 11 hours on a single V100 GPU with 9.5GB memory. Regularization terms, $\lambda_O$ and $\lambda_{\text{pen}}$, are assigned values of 5 and 3, respectively. 
During the contact optimization, we apply the random restart to avoid a local optimum following~\cite{grady2021contactopt}. NL2Contact takes around 3 seconds to generate per contact map.

\subsection{Datasets}
\noindent\textbf{\textit{ContactDescribe} dataset.} The \textit{ContactDescribe} dataset is an enhanced dataset based on the ContactPose. ContactPose is the first dataset~\cite{brahmbhatt2020contactpose} that combines hand-object contact information with hand pose, object pose, and RGB-D images. It encompasses 2,306 distinct grasps involving 25 household objects, executed with two functional intents by 50 participants, and includes over 2.9 million RGB-D grasp images. 
To ensure fair comparisons with the ContactOpt~\cite{grady2021contactopt} and S$^2$Contact~\cite{tse2022s}, we adhere to the Perturbed ContactPose dataset, wherein hand meshes undergo modification through the introduction of additional noise to MANO parameters. This dataset comprises 22,624 training grasps and 1,416 testing grasps. We annotated five contact descriptions for each grasp instance. Finally, we train our NL2Contact on the \textit{ContactDescribe} dataset. 

\noindent\textbf{Described HO3D dataset.}
HO3D~\cite{hampali2020honnotate} datasets are used for evaluating 3D hand-object interactions, which are captured in a real-world setting and involve 10 different subjects performing various fine-grained manipulations on one of the 10 objects from the YCB models~\cite{calli2015ycb}. We employ the HO3D dataset to assess the generalization capability of our proposed method when faced with previously unseen objects. We choose a subset from HO3D following ContactOpt~\cite{grady2021contactopt} (around 14k grasps) and manually annotate the textual descriptions.

\begin{figure*}[t]
\centering
\includegraphics[width=0.95\linewidth]{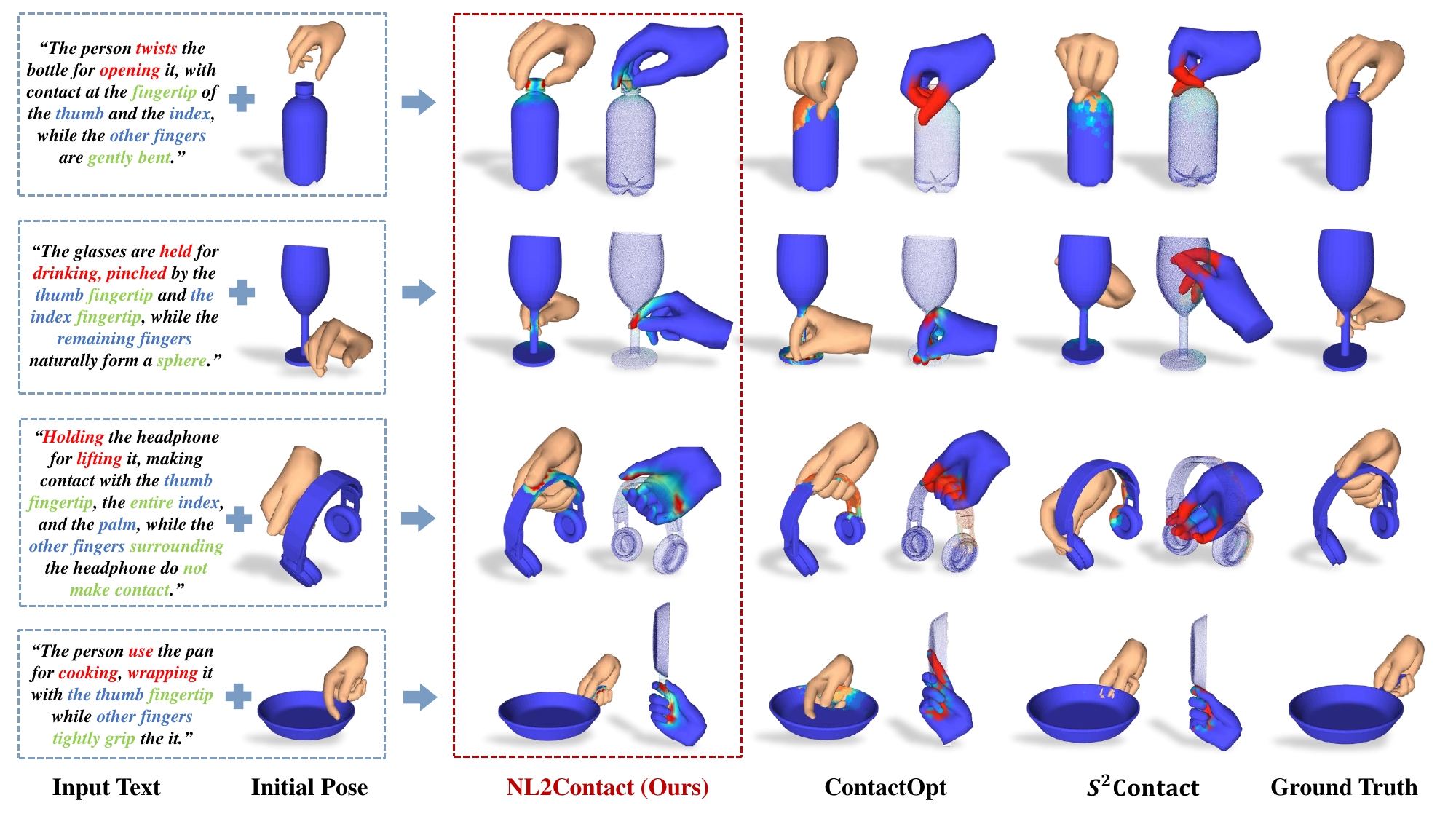}
\caption{
We conduct qualitative comparisons with state-of-the-art methods.
Our method generates accurate and controllable contact, matching the input text description.
ContactOpt~\cite{grady2021contactopt} and S$^2$Contact~\cite{tse2022s} are uncontrollable contact generation methods. They tend to generate a grasp with all five fingers engaged. Interestingly, we provide a great initial pose in the first row.
However, ContactOpt and S$^2$Contact still generate a five-figner grasp which is different from the initial pose.
}
\label{fig:Qulitative}
\end{figure*}

\subsection{Evaluation Metrics}
\noindent\textbf{Hand error.} We use the Mean Per-Joint Position Error (MPJPE) ($\mathrm{mm}$) over 21 joints to evaluate the grasp refinement. We calculate the average L2 per-joint kinematic error with the ground truth hand pose.

\noindent\textbf{Contact quality.}
We adopt the intersection volume ($\mathrm{cm}^3$), contact coverage ($\%$) and 
 Contact Precision/Recall ($\%$). The intersection volume (denoted as
Inter. in tables) is acquired by voxelizing both the hand and object with a voxel size of 0.5cm. 
Contact coverage (denoted as
Cover. in tables) calculates the proportion of grasps in contact with objects, defined as the percentage of hand points within ±2mm of the object's surface. 
Contact precision/recall (denoted as
Pr. and Re. in tables) calculates the similarity between
the generated contact and ground-truth contact with a threshold of 0.4. 

\noindent\textbf{Diversity of grasp generation.} We cluster generated grasps into 20 clusters using K-means and then evaluated the average Cluster Size (denoted as Diversity).

\noindent\textbf{Physical stability.} We placed the object and its corresponding grasp in a simulator. We evaluated the simulation displacement (SD) by calculating the displacement (cm) of the object's center relative to its center of mass over a specified duration under the influence of gravity.
We randomly generated 20 grasps and reported the average SD.

\subsection{Evaluating Grasp Pose Optimization}
We first evaluate the effectiveness of the proposed NL2Contact application on grasp pose optimization from the hand-object point cloud.

\noindent\textbf{Baseline methods.}
We compare our method with three strong
geometry-guided grasp optimization baselines. 1) ContactOpt~\cite{grady2021contactopt} estimates contact maps on both hand and object with PointNet-based DeepContact.  2) S$^2$Contact~\cite{tse2022s} proposed a graph-based GCN-Contact network to predict hand-object contacts and a semi-supervised method to improve generalization on out-of-domain datasets. 3) Jiang~\etal~\cite{jiang2021hand} also provides a ContactNet to estimate the contact map from the hand-object point cloud, but it only outputs the object contact.

\begin{table}[t]
\centering
        \small
	\caption{\textbf{Quantitative results of grasp pose optimization on \textit{ContactDescribe} and Described HO3D dataset.} Due to the natural language guidance, we can model accurate hand-object contact. The optimization from our predicted contact achieves the lowest MPJPE. We also show that contact descriptions improve the generalization of our method to out-of-domain objects on HO3D dataset.}
        \setlength\tabcolsep{3pt} 
        \resizebox{\linewidth}{!}{
	\begin{tabular}{lccccc|ccc}
		\toprule[1.2pt]
              \multicolumn{6}{c|}{\textbf{\textit{ContactDescribe} Dataset}}
             &\multicolumn{3}{c}{\textbf{Described HO3D Dataset}} \\
            \multirow{2}{*}{\textbf{Method}} &\multirow{2}{*}{\textbf{MPJPE}$\downarrow$} & \multicolumn{4}{c|}{\textbf{Contact Quality}} &\multirow{2}{*}{\textbf{Method}} &\multirow{2}{*}{\textbf{MPJPE ($mm$)$\downarrow$}} &\multirow{2}{*}{\textbf{Inter.} ($mm^3$)$\downarrow$}\\
            \cline{3-6}
		& &\textbf{Inter.} $\downarrow$ &\textbf{Cover.} $\uparrow$ &\textbf{Pr.} $\uparrow$ &\textbf{Re.} $\uparrow$ \\
		\hline
            Perturbed Pose& 79.9	& 8.4&	$2.3\%$	&$9.9\%$&$	11.5\%$ &Initial Pose~\cite{hasson20_handobjectconsist} & 11.4	& 9.26\\
            ContactNet~\cite{jiang2021hand}& 45.2	& 15.6&	$18.4\%$&	$31.6\%$&	$47.6\%$ &ContactOpt~\cite{grady2021contactopt} & 9.5	& 5.71\\
            ContactOpt~\cite{grady2021contactopt} & 25.1 & 12.8  & $19.7\%$ & $38.7\%$&$54.8\%$ &TOCH~\cite{zhou2022toch}& 9.3	& 4.67\\
            S$^2$Contact~\cite{tse2022s} &29.4	&12.2	&$22.2\%$	&$42.5\%$	&$56.1\%$ &S$^2$Contact~\cite{tse2022s}& 8.7	& \textbf{3.52}\\
            \rowcolor[rgb]{0.902,0.902,0.902} NL2Contact & \textbf{21.7}	&  \textbf{7.1} & $\textbf{30.5}\%$ & $\textbf{49.2}\%$&$\textbf{59.9}\%$ &NL2Contact & \textbf{8.4}	&  4.39\\
		\bottomrule[1.2pt]
	\end{tabular}}
	\label{tab:result_opt}
\end{table}

\noindent\textbf{Results on \textit{ContactDescribe}.}
We evaluate the NL2Contact on the ContactDescribe dataset, in which we perturb the annotated MANO parameters. The hand error after perturbation is around 80 mm. This tests the ability to improve hand poses using the predicted contact maps.
Table~\ref{tab:result_opt} shows a quantitative comparison between different methods. Our method significantly outperforms the
other methods by a large margin. 
We achieve the lowest MPJPE score, outperforming ContactOpt~\cite{grady2021contactopt} by 4.4 $mm$. Besides, our method significantly reduced the intersection volumes while other methods increased it. As shown in Figure~\ref{fig:Qulitative}, the ContactOpt and S$^2$Contact always have penetration cases. For the contact quality, the results illustrate that our method can generate more similar contact with the ground-truth contact (gathering by thermal camera). We can observe that the NL2Contact is the most effective one for hand pose refinement. Other methods only boost the contact quality, and the hand pose still exhibits large errors after contact optimization. However, our method leverages the language description to provide semantic information for generating the contact map and effectively reduces hand error while improving the contact quality.

\noindent\textbf{Results on Described HO3D.} We demonstrate the generalization ability
of our method on the HO3D dataset. Since the hand model (MANO) is consistent across different datasets, most contact modeling methods can directly infer the contact without being retrained. Nevertheless, the domain gap still exists, occasionally resulting in insufficient contact or significant penetration. 
As indicated in Table~\ref{tab:result_opt}, our method's performance is close to the best method, S$^2$Contact~\cite{tse2022s}, in the metric of Intersection Volume. The S$^2$Contact used a complex semi-supervised pipeline and carefully selected pseudo-label to model contact for an unseen object. This illustrates that our designed hand-centered contact description is generic and adapts to the large domain among different datasets. Besides, compared with S$^2$Contact, our method achieves notably lower hand error.

\noindent\textbf{Qualitative results.}
We present our main experimental results in Figure~\ref{fig:Qulitative}.
We observe that our proposed method can generate accurate contacts that exactly match the input concrete descriptions, and then the initial hand pose can be improved by contact optimization. Other methods, e.g. ContactOpt~\cite{grady2021contactopt} and S$^2$Contact~\cite{tse2022s} always generate full-five finger grasps. Besides, their method is not controllable; results after refinement may exhibit larger hand pose errors, while our method, guided by textual prompts, ensures that the optimized results align with the target hand pose.

\begin{table}[t]
\centering
  \begin{minipage}{0.48\linewidth}
        \small
	\caption{\textbf{Quantitative results of grasp generation.} Our method achieves the lowest penetration, best contact, and highest diversity compared to ContactGen~\cite{liu2023contactgen}.}
    \resizebox{\linewidth}{!}{
	\begin{tabular}{lcccc}
		\toprule[1.2pt]
            \textbf{Method} &\textbf{Inter.}$\downarrow$  &\textbf{Conver.}$\uparrow$
            &\textbf{Diversity}$\uparrow$
            &\textbf{SD}$\downarrow$\\
		\hline
    GrabNet~\cite{taheri2020grab} & 15.50	&\textbf{99}$\%$ & 2.06 &2.34 \\
    GraspTTA~\cite{jiang2021hand} & 7.37	&$76\%$ &	1.43 &5.34\\
    ContactGen~\cite{liu2023contactgen} & 9.96	&$97\%$ &	5.04 &2.70\\
    \rowcolor[rgb]{0.902,0.902,0.902} Ours & \textbf{5.89} &\textbf{99}$\%$ &\textbf{5.91} &\textbf{2.31}\\
		\bottomrule[1.2pt]

 \end{tabular}}
	\label{tab:result_generation}
\end{minipage}
~  
\begin{minipage}{0.48\linewidth}
\centering
\includegraphics[width=1.0\linewidth]{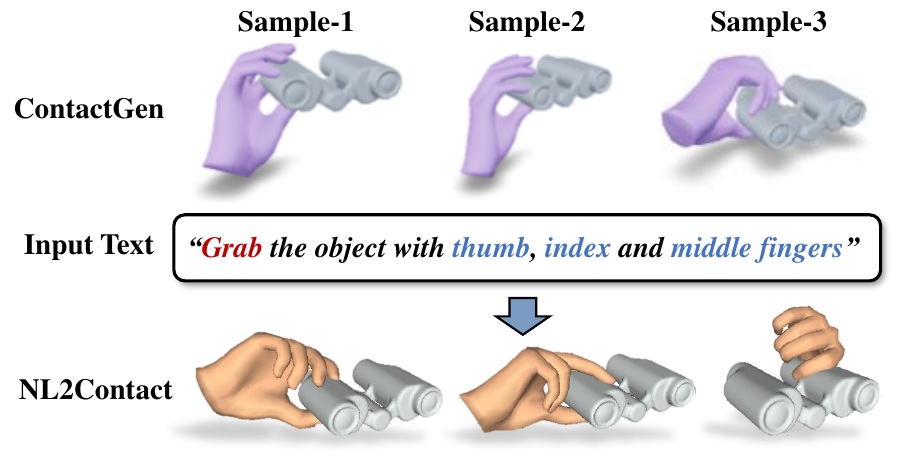}
\captionof{figure}{
\textbf{Visualization of diverse grasp generation.} We
observe ContactGen~\cite{liu2023contactgen} always generates grasps with all five fingers engaged, while our method generates grasps consistent with the text.
}
\label{fig:vis_generation}
\end{minipage}
\end{table}

\subsection{Evaluating Human Grasp Generation}
Our method can be extended to generate human grasps from a given input object and contact description. We train the model on the \textit{ContactDescribe} dataset and test on unseen objects from the HO3D dataset.

\noindent\textbf{Baseline methods.}
We compare our method with the state-of-the-art human grasp generation approach, ContactGen~\cite{liu2023contactgen}, which generates contact for the input object and then adopts model-based optimization to predict grasps.

\noindent\textbf{Results.} 
We first quantitatively evaluate the grasp generation for the unseen object from the HO3D dataset. GraspTTA~\cite{jiang2021hand} are trained on the ObMan dataset~\cite{hasson2019learning}, and GrabNet~\cite{taheri2020grab} and ContactGen~\cite{liu2023contactgen} are trained on the GRAB dataset~\cite{taheri2020grab}. We randomly generate 20 grasps. As indicated in Table~\ref{tab:result_generation}, our method achieves the best performance in all metrics. GraspTTA and GrabNet are struggling with poor diversity. Comparison results of grasp generation for binoculars are shown in Figure~\ref{fig:vis_generation}. ContactGen always generates a grasp with all five fingers touched, while our method can control the generation using the text.

\subsection{Ablation Study}
\noindent\textbf{Effect of multi-level text input.} To validate the effectiveness of the different level text descriptions, we conducted the experiments by varying the specificity of the text and showing the grasp optimization results in Table~\ref{tab:ab_text}. The initial grasp pose is also from Hasson~\etal~\cite{hasson20_handobjectconsist}.
We find that the more \textit{low-level} text, the more precise the grasp and location of the contact points.
Notably, the high-level text is removed from all contact-related descriptions and still significantly outperforms ContactOpt. 

\begin{table}[t]
\centering
  \begin{minipage}{0.42\linewidth}
        \small
	\caption{\textbf{Ablation study of multi-level text input.} We demonstrate
that the more low-level, the more precise the grasp and location of the contact points.}
        \resizebox{\linewidth}{!}{
	\begin{tabular}{lcccc}
		\toprule[1.2pt]
            \multirow{2}{*}{\textbf{Method}} &\multirow{2}{*}{\textbf{MPJPE}$\downarrow$} & \multicolumn{3}{c}{\textbf{Contact Quality}}\\
            \cline{3-5}
		& &\textbf{Inter.} $\downarrow$ &\textbf{Pr.} $\uparrow$ &\textbf{Re.} $\uparrow$ \\
		\hline
            ContactOpt~\cite{grady2021contactopt} & 25.1	&12.8 &	$38.7\%$&	$54.8\%$\\
            High-level & 23.9	& 8.9	&$42.7\%$&$	55.6\%$\\
            Middle-level & 22.5 & 8.0  & $46.8\%$&$57.3\%$\\
            \rowcolor[rgb]{0.902,0.902,0.902} \textbf{Low-level} & \textbf{21.7}	&  \textbf{7.1} & $\textbf{49.2}\%$&$\textbf{59.9}\%$\\
		\bottomrule[1.2pt]
	\end{tabular}}
	\label{tab:ab_text}
\end{minipage}
~  
\begin{minipage}{0.54\linewidth}
\centering
        \small
\caption{\textbf{Ablation study of our method.} We demonstrate
that the full model of NL2Contact outperforms all other variants.}
        \vspace{-2mm}
        \resizebox{\linewidth}{!}{
	\begin{tabular}{lcccc}
		\toprule[1.2pt]
            \multirow{2}{*}{\textbf{Method}} &\multirow{2}{*}{\textbf{MPJPE}$\downarrow$} & \multicolumn{3}{c}{\textbf{Contact Quality}}\\
            \cline{3-5}
		& &\textbf{Inter.} $\downarrow$ &\textbf{Pr.} $\uparrow$ &\textbf{Re.} $\uparrow$ \\
		\hline
            w/o text-guided & 24.3	& 11.7	&$39.9\%$&$	56.2\%$\\
            \hline
            w/o text2hand fusion & 23.5 & 10.3  & $41.6\%$&$57.9\%$\\
            w/o text2ho fusion  & 22.9	& 9.7&	$43.9\%$&	$58.4\%$\\
            \hline
            w/o staged diffusion & 27.3	& 13.6&	$38.5\%$&	$52.1\%$\\
            \hline
            with VAE &21.9	&7.4	&$47.0\%$	&$56.3\%$\\
            \hline
            \rowcolor[rgb]{0.902,0.902,0.902} \textbf{full model} & \textbf{21.7}	&  \textbf{7.1} & $\textbf{49.2}\%$&$\textbf{59.9}\%$\\

		\bottomrule[1.2pt]
	\end{tabular}}
	\label{tab:result_ab}
\end{minipage}
\end{table}

\noindent\textbf{Effect of language description.}
To validate the effectiveness of the language description-guided contact modeling, we conduct the experiments with the following settings:
(a) Without text input: the contact description is not used. At the Text-to-Hand-Object fusion stage, we use the multi-head attention 2 module to fuse the object global feature with the hand pose feature to produce conditional embedding. At the text-ho stage, we do not concatenate the text global feature.
(b) Without Text-to-Hand fusion module: the text feature does not fuse to the hand pose feature using cascaded multi-head attention, instead by directly concatenating.
(c) Without the Text-to-HO fusion module, the text feature does not fuse with the hand-object point cloud feature while fusing the global hand-object feature into the local hand-object feature.
The results of the ablation study are shown in Table~\ref{tab:result_ab}. Comparing (a) with the full model, we find the hand error significantly increases while the contact quality decreases.  This result verifies that introducing contact description supervision improves the effectiveness of our model. However, our results are still better than those of ContactOpt; we consider the reason that our diffusion model is more effective than PointNet for modeling contact. Comparing (b) and (c), we find that Text-to-Hand fusion is more effective because this can correct the initial input to fit the gesture in the description and reduce the complexity of text-to-contact generation.

\noindent\textbf{Effect of the Staged Diffusion model.}
To validate the effectiveness of the staged diffusion, we conduct the experiments with the following settings:
(d) Without the staged diffusion model: we do not generate hand-object contact with two stages. Instead, we use a single diffusion model to generate the contact map directly. To this end, we use PointNet to extract the hand feature from the initial hand pose and the Text-to-HO fusion module to fuse all features.
(e) With VAE: we implement our pipeline using two VAE generators instead of staged diffusion.
As illustrated in Table~\ref{tab:result_ab}, the performance declines significantly under setting (d) without generated hand pose.  This result demonstrates that directly cross-modal mapping text to hand-object contact is non-trivial. Setting (e) also shows that other generative models, such as VAEs, apply to our task. However, the diffusion model performs better due to its iterative denoising mechanism, which is more effective in learning a data distribution.

\begin{figure*}[t]
\centering
\includegraphics[width=1\linewidth]{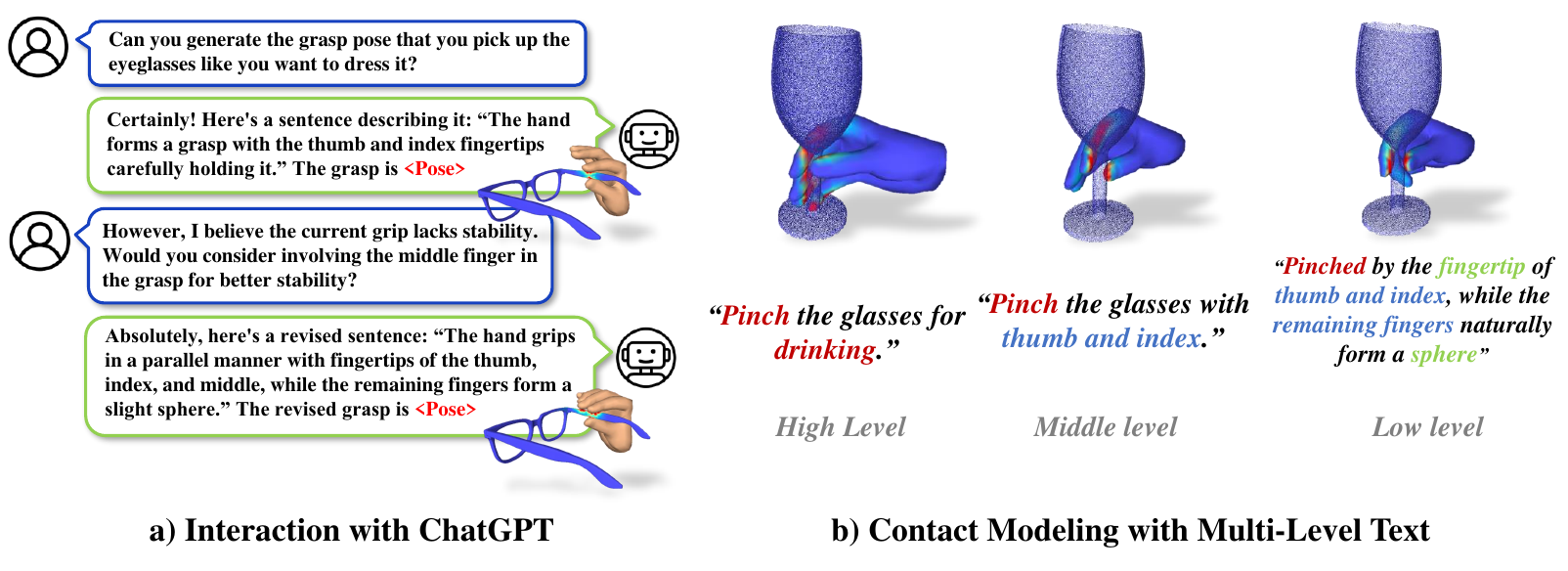}
\caption{
a) We show the seamless integration of our method with ChatGPT, allowing for interactive control to generate an initial grasp and refine it. Users offer high-level prompts, and ChatGPT generates detailed contact descriptions, then feeds descriptions into our network to generate the grasp. b) 
Our model generates more accurate grasps as text descriptions become more detailed.
}
\label{fig:chatgpt_ctl}
\end{figure*}

\noindent\textbf{Applications with ChatGPT.}
Our method can provide more potential downstream applications, such as the teleoperation of robotic hands and virtual avatar editing. In this experiment,
we integrate NL2Contact with ChatGPT~\cite{chatgpt} to facilitate interactive control.
As illustrated in the left part of Figure~\ref{fig:chatgpt_ctl}, the user initially generates a grasp of the scissors using a high-level description. The large language model comprehends how to use the scissors and provides a low-level contact description. Subsequently, our method generates the initial grasp by low-level description. If the generated contact is unsatisfactory, our method can optimize the grasp pose based on different levels of user feedback, as shown in the right part of Figure~\ref{fig:chatgpt_ctl}. 

\section{Conclusion}
This paper proposes a novel task - natural language-guided 3D hand-object contact modeling.
We introduce NL2Contact, a novel cross-modal latent diffusion model for contact modeling,
and \textit{ContactDescribe} dataset, which provides 3D hand-object contact and fine-grained natural language descriptions.
Our method can be applied to grasp pose optimization and novel human grasp generation.  
In the future, we would like to explore the modeling of dynamic contacts with text.

\clearpage  

\section*{Acknowledgements}
This research was supported by the MSIT(Ministry of Science and ICT), Korea, under the ITRC(Information Technology Research Center) support program(IITP-2024-2020-0-01789) supervised by the IITP(Institute for Information \& Communications Technology Planning \& Evaluation), the National Research Foundation Singapore and DSO National Laboratories under the AI Singapore Programme (Award Number: AISG2-RP-2020-016), China Scholarship Council (CSC) Grant No. 202208060266 and No.202006210057.

%
%
\bibliographystyle{splncs04}
\bibliography{main}
\end{document}